\documentclass[letterpaper]{article}

\usepackage{wrapfig}
\usepackage[header]{aimatters}
\usepackage{apacite}
\usepackage{graphicx}
\usepackage{hyperref}
\usepackage{url}

\title{Artificial Inteligence Buzzword Explained: \\ Multi-Agent Path Finding
  (MAPF)\thanks{ This article is a service of the ACM Special Interest Group
    on Artificial Intelligence (ACM SIGAI). If you want to find out more about
    artificial intelligence and are not a member of ACM SIGAI yet, you can
    join for \$11 (students) or \$25 (others), see \url{sigai.acm.org}
    (membership fees are accurate for 2016).}}  \author{ \AIMauthor{Hang
    Ma}{University of Southern
    California}{hangma@usc.edu} 
  \AIMauthor{Sven Koenig}{University of Southern California}{skoenig@usc.edu}
}

\AIMbio{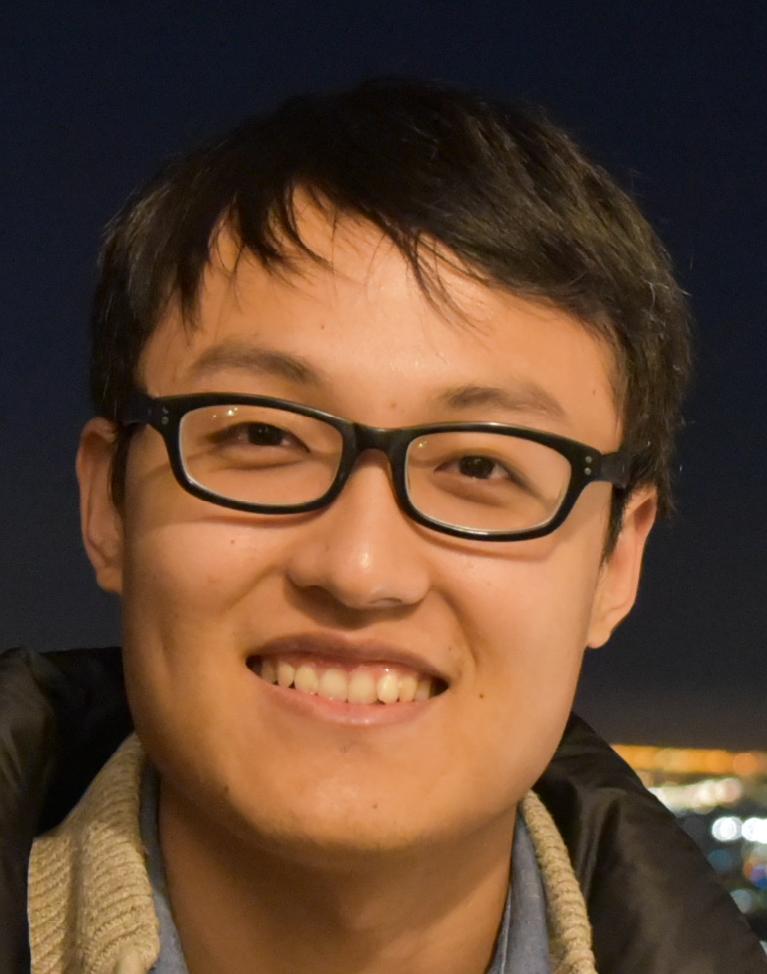}{\textbf{Hang Ma} is a Ph.D. student in computer science at
  the University of Southern California. He is the recipient of a USC
  Annenberg Graduate Fellowship. His research interests include artificial
  intelligence, machine learning and robotics. Additional information about
  Hang can be found on his webpages:
  \url{http://www-scf.usc.edu/~hangma}. Contact him at
  \href{mailto:hangma@usc.edu}{hangma@usc.edu}.}

\AIMbio{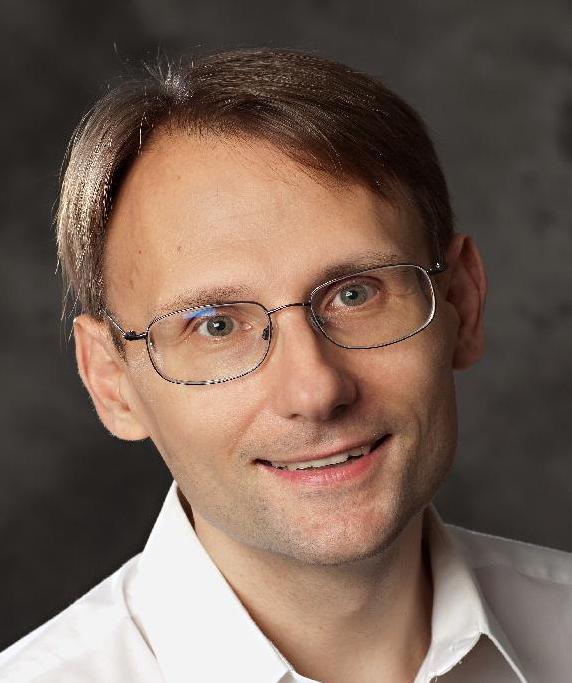}{\textbf{Sven Koenig} is a professor in computer science at
  the University of Southern California. Most of his research centers around
  techniques for decision making (planning and learning) that enable single
  situated agents (such as robots) and teams of
  agents to act intelligently in their environments and exhibit goal-directed
  behavior in real-time. Additional information about
  Sven can be found on his webpages: \url{http://idm-lab.org}. Contact him at
  \href{mailto:skoenig@usc.edu}{skoenig@usc.edu}.}

\begin{document}
\maketitle

\begin{figure*}[th]
\center
\includegraphics[height=50pt]{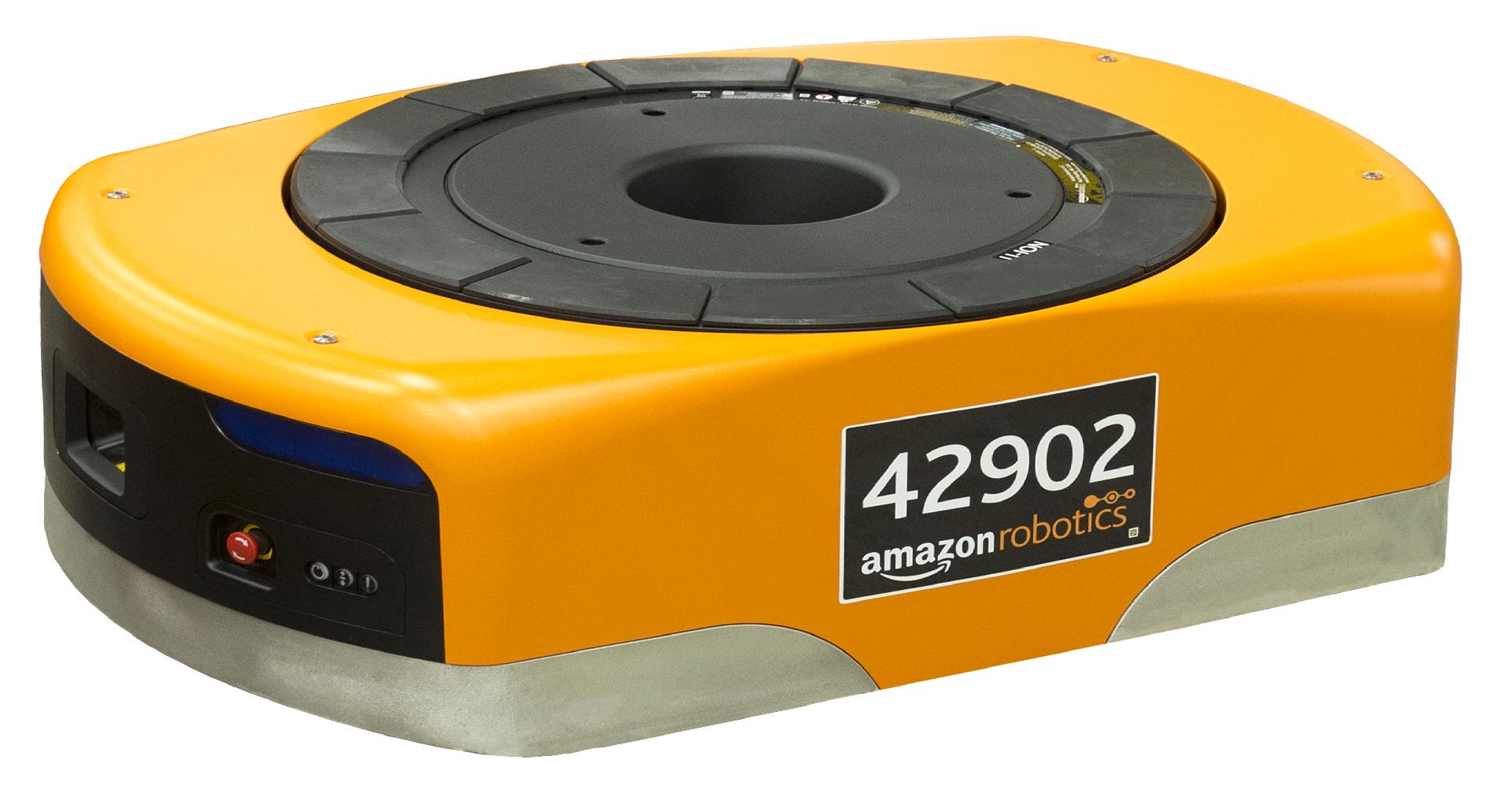}
\includegraphics[height=100pt]{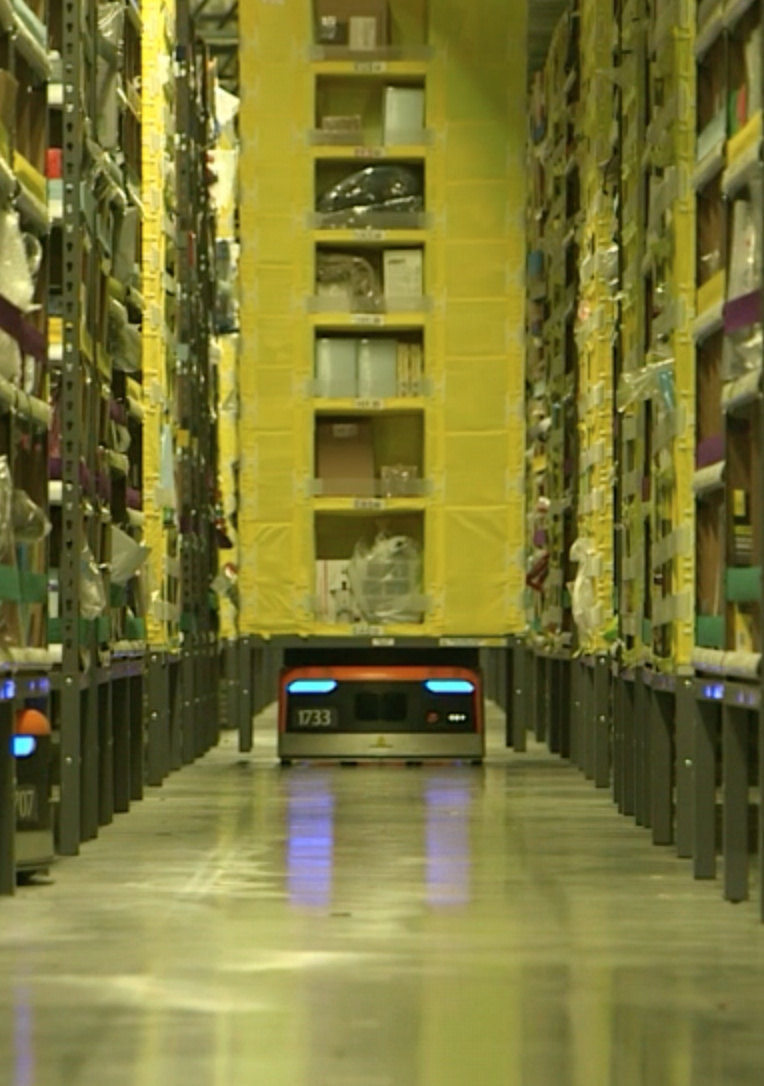} \hspace*{0.5mm}
\includegraphics[height=100pt]{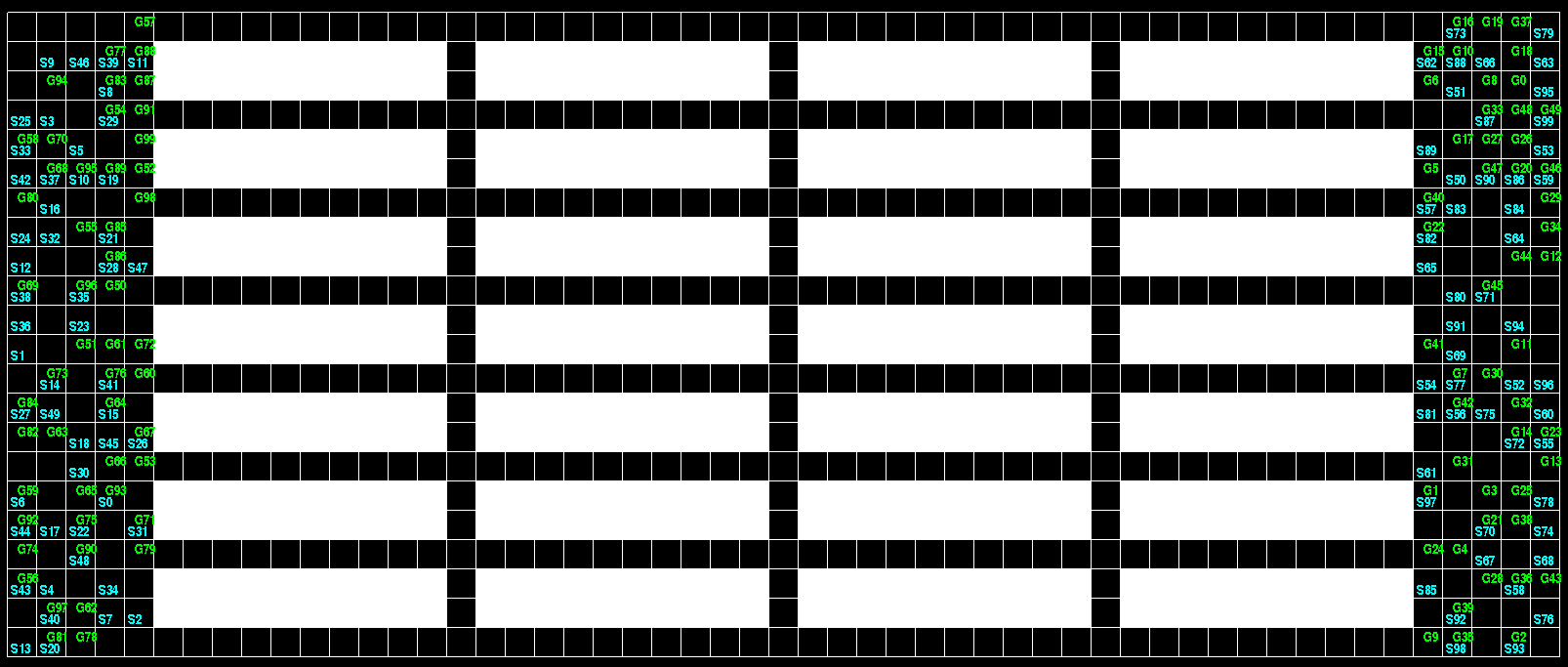}
\caption{Warehouse robot (left), inventory pods (center), and the layout of a
  small simulated warehouse (right). The left and center photos are courtesy
  of Amazon Robotics.}
  \label{fig:kiva}
\end{figure*}

Kiva Systems was founded in 2003 to develop robot technology that automates
the fetching of goods in order-fulfillment centers. It was acquired by Amazon
in 2012 and changed its name to Amazon Robotics in 2014. Amazon
order-fulfillment centers have inventory stations on the perimeter of the
warehouse and storage locations in its center, see Figure~\ref{fig:kiva}. Each
storage location can store one inventory pod. Each inventory pod holds one or
more kinds of goods. A large number of warehouse robots operate autonomously
in the warehouse. Each warehouse robot is able to pick up, carry and put down
one inventory pod at a time. The warehouse robots move inventory pods from
their storage locations to the inventory stations where the needed goods are
removed from the inventory pods (to be boxed and eventually shipped to
customers) and then back to the same or different empty storage
locations \cite{kiva}.\footnote{See the following YouTube video: \\
  \url{https://www.youtube.com/watch?v=6KRjuuEVEZs}}

\begin{figure*}[th]
\center
\includegraphics[width=\textwidth]{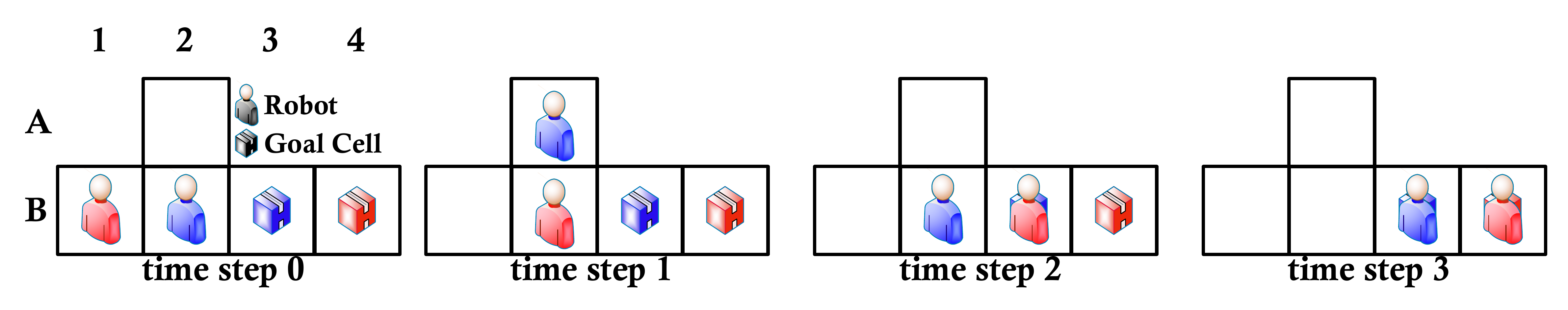}
\caption{A multi-agent path-finding instance with two robots.}
\label{fig:MAPF}
\end{figure*}

These order-fulfillment centers raise a number of interesting optimization
problems, such as which paths the robots should take and at which storage
locations inventory pods should be stored. Path planning, for example, is
tricky since most warehouse space is used for storage locations, resulting in
narrow corridors where robots that carry inventory pods cannot pass each
other. Warehouse robots operate all day long but a simplified one-shot version
of the path-planning problem is the multi-agent path-finding (MAPF) problem,
which can be described as follows: On math paper, some cells are blocked. The
blocked cells and the current cells of $n$ robots are known. A different
unblocked cell is assigned to each of the $n$ robots as its goal cell. The
problem is to move the robots from their current cells to their goal cells in
discrete time steps and let them wait there. The optimization objective is to
minimize the makespan, that is, the number of time steps until all robots are
at their goal cells.  During each time step, each robot can move from its
current cell to its current cell (that is, wait in its current cell) or to an
unblocked neighboring cell in one of the four main compass directions. Robots
are not allowed to collide. Two robots collide if and only if, during the same
time step, they both move to the same cell or both move to the current cell of
the other robot.  Figure~\ref{fig:MAPF} shows an example, where the red and
blue robots have to move to the red and blue goal cells, respectively.

\begin{figure}[th]
\center
\includegraphics[height=60pt]{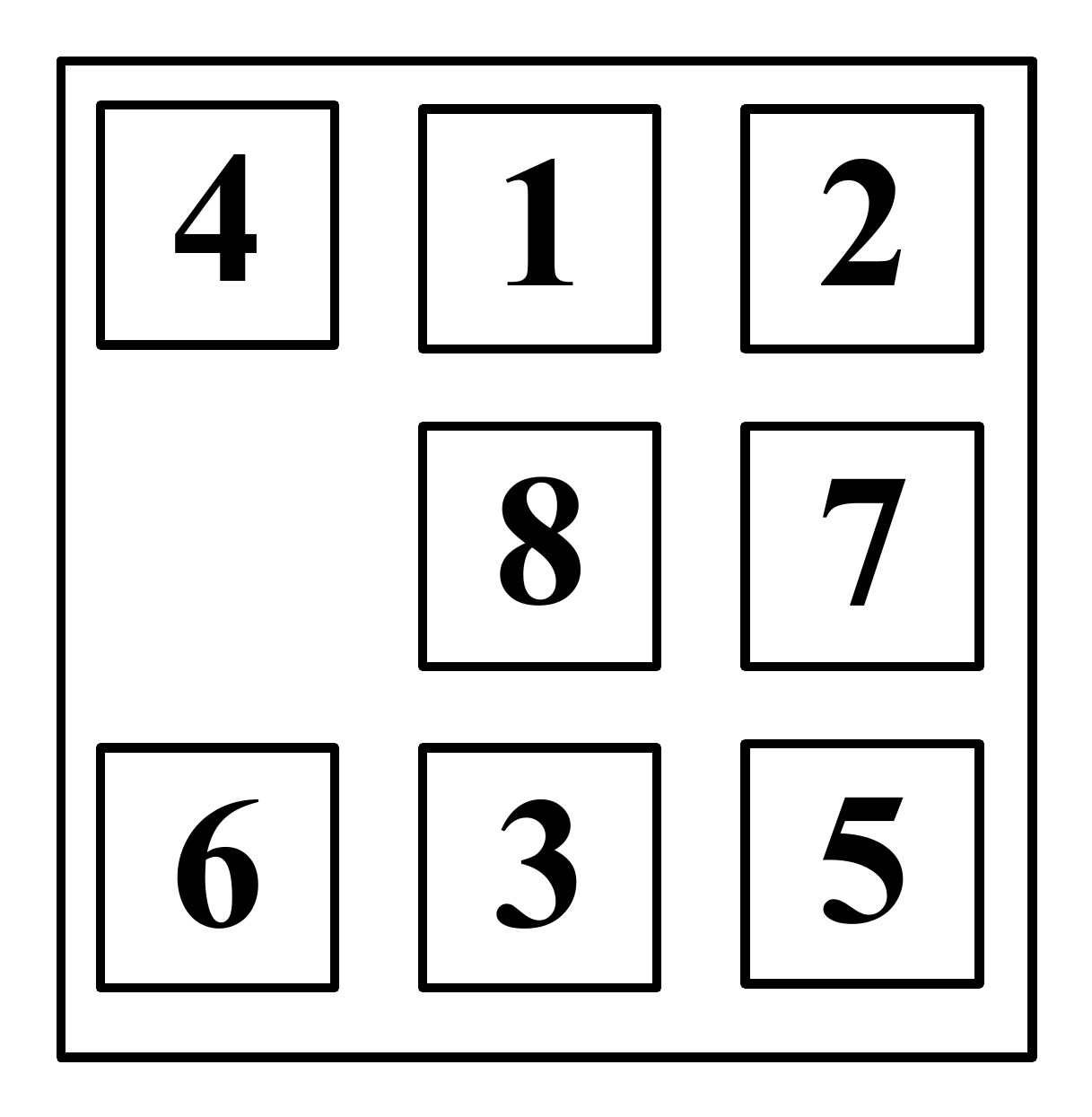}
\caption{The eight-puzzle.}
\label{fig:puzzle}
\end{figure}

There are also versions of the multi-agent path-finding problem with different
optimization objectives than makespan (such as the sum of the time steps of
each robot until it is at its goal cell) or slightly different collision or
movement rules. For example, solving the eight-puzzle (a toy with eight square
tiles in a three by three frame, see Figure~\ref{fig:puzzle}) is a version of
the multi-agent path-finding problem where the tiles are the robots.

Researchers in theoretical computer science, artificial intelligence and
robotics have studied multi-agent path finding under slightly different
names. They have developed fast (polynomial-time) algorithms that find
solutions for different classes of multi-agent path-finding instances (for
example, those with at least two unblocked cells not occupied by robots)
although not necessarily with good makespans. They have also characterized the
complexity of finding optimal (or bounded-suboptimal) solutions and developed
algorithms that find them. A bounded-suboptimal solution is one whose makespan
is at most a given percentage larger than optimal.

\begin{figure*}[th]
\center
\includegraphics[width=0.9\textwidth]{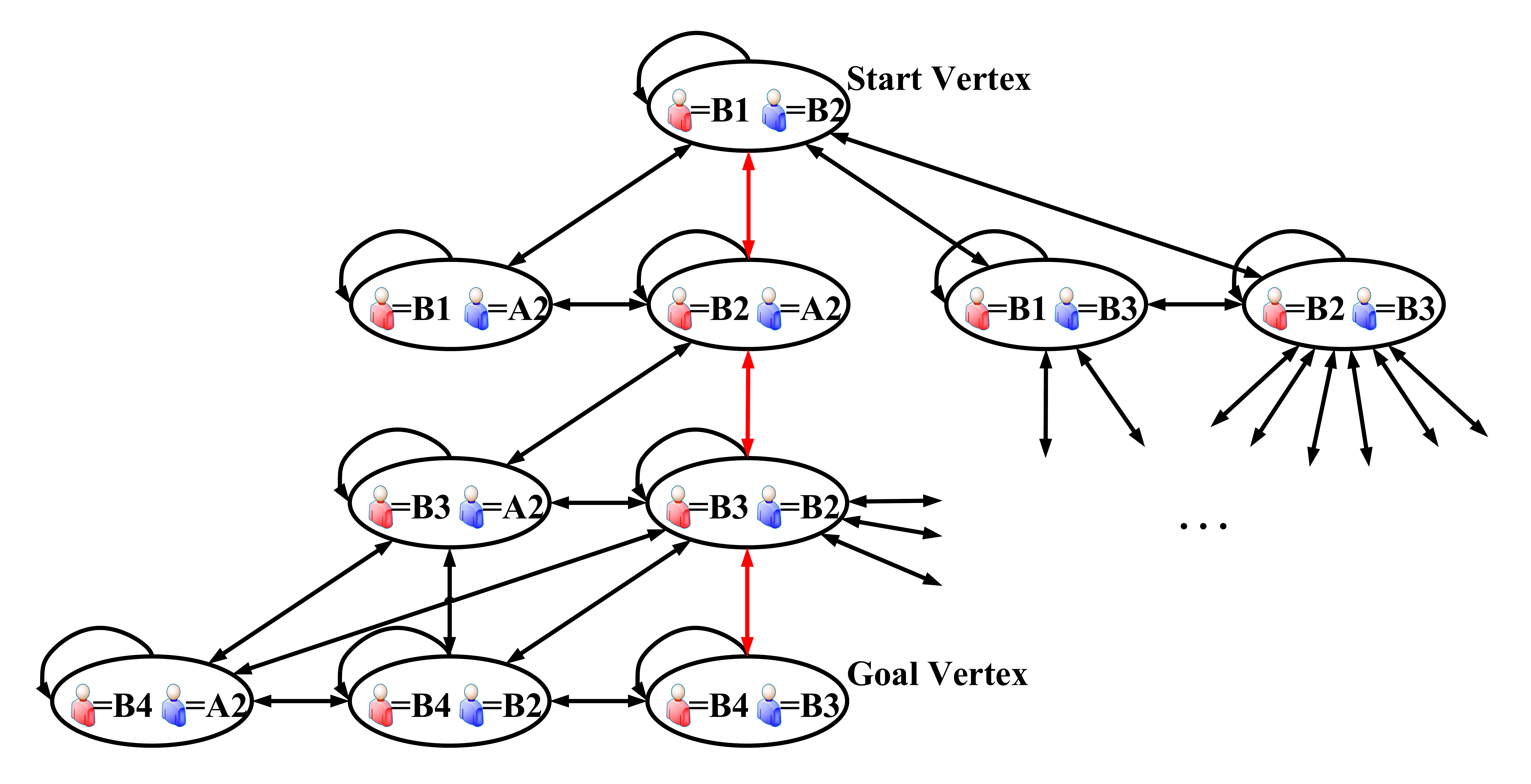}
\caption{Partial graph for the multi-agent path-finding instance from Figure \ref{fig:MAPF}.}
\label{fig:graph}
\end{figure*}

Interestingly, it is slow (NP-hard) to find optimal
solutions~\cite{YuLav13AAAI,MaAAAI16}, although a slight modification of the
multi-agent path-finding problem can be solved in polynomial time with flow
algorithms, namely where $n$ unblocked cells are given as goal cells but it is
up to the algorithm to assign a different goal cell to each one of the $n$
robots~\cite{YuLav13STAR}. Researchers have also studied versions of the
multi-agent path-finding problem where goal cells require robots with certain
capabilities \cite{MaAAMAS16} or robots can exchange their payloads
\cite{MaAAAI16}.

In principle, one can model the original multi-agent path-finding problem as a
shortest-path problem on a graph whose vertices correspond to tuples of cells,
namely one for each robot, as shown in Figure \ref{fig:graph} (where the red
path shows the optimal solution), but the number of vertices can be
exponential in the number of robots and the shortest path thus cannot be found
quickly. Instead, researchers have suggested to plan a shortest path for each
robot independently (by ignoring the other robots), which can be done fast. If
all robots can follow their paths without colliding, then an optimal solution
has been found. If not, then ...

\begin{itemize}

\item there are multi-agent path-finding algorithms that group all colliding
  robots together and find a solution for the group with minimal makespan (by
  ignoring the other robots), and then repeat the process.  The hope is to
  find a solution before all robots have been grouped together into one big
  group~\cite{ODA,ODA11}.

\item there are other multi-agent path-finding algorithms that pick a
  collision between two robots (for example, robots A and B both move to cell
  x at time step t) and then consider recursively two cases, namely one where
  robot A is not allowed to move to cell x at time step t and one where robot
  B is not allowed to move to cell x at time step t. The hope is to find a
  solution before all possible constraints have been
  imposed~\cite{DBLP:journals/ai/SharonSFS15}.

\end{itemize}

These state-of-the-art multi-agent path-finding algorithms are currently not
quite able to find bounded-suboptimal solutions for 100 robots in small
warehouses in real-time. The tighter the space, the longer the runtime.
Researchers have also suggested a variety of other multi-agent path-finding
techniques~\cite{WHCA,WHCA06,Ryan08,WangB08,WangB11,PushAndSwap,DBLP:journals/ai/SharonSGF13,PushAndRotate,ECBS,EPEJAIR,MStar,ICBS,MaAAMAS16,CohenUK16},
including some that transform the problem into a different problem for which
good solvers exist, such as satisfiability~\cite{Surynek15}, integer linear
programming~\cite{YuLav13ICRA} and answer set
programming~\cite{erdem2013general}. Researchers have also studied how to
execute the resulting solutions on actual robots \cite{Koen14b,Koen16g}.

Two workshops have recently been held on the topic, namely the AAAI 2012
Workshop on Multi-Agent Pathfinding\footnote{See the following URL: \\
  \url{http://movingai.com/mapf}} and the IJCAI 2016 Workshop on Multi-Agent
Path Finding\footnote{See the following URL: \\
  \url{http://www.andrew.cmu.edu/user/gswagner/workshop/ijcai\_2016\_multirobot\_path\_finding.html}}.
Recent dissertations include~\cite{Wang12,Wagner15,Sharonxx}.

\bibliography{references}
\bibliographystyle{apacite}

\end{document}